\documentclass[12pt]{article}

\usepackage[a4paper,margin=1in]{geometry}
\usepackage{amssymb}
\usepackage{amsmath}
\usepackage{float}
\usepackage[colorlinks=true, linkcolor=blue, citecolor=red, urlcolor=magenta]{hyperref}
\usepackage{tikz}
\usepackage{graphicx}
\usepackage{xcolor}
\usepackage{authblk}
\usepackage{natbib}
\usepackage{orcidlink}
\setcitestyle{numbers,square,comma}

\title{Adapting Visualization Techniques for Time-Series Anomaly Detection: From Convolutional Neural Networks to Convolutional–Recurrent Neural Networks}

\author[1]{Fabien Poirier\orcidlink{0000-0002-2203-7571}}
\author[1,2]{Myriam Lamolle\orcidlink{0000-0001-9652-7891}}
\affil[1]{Université Paris 8 - IUT de Montreuil, LIASD, France}
\affil[2]{Centre de Génie Industriel (CGI), IMT Mines Albi, Albi, France}

\date{}

\begin{document}

\maketitle

\begin{abstract}
Deep neural networks achieve strong performance on complex tasks but are often regarded as “black boxes,” which limits their adoption in domains where transparency is essential. This lack of interpretability raises ethical and legal concerns, particularly in sensitive applications such as security, where automated decisions can have serious consequences. The General Data Protection Regulation (GDPR) reinforces the need to justify decisions made by these systems. In this work, we investigate visualization techniques to improve the interpretability of anomaly detection models based on convolutional recurrent neural networks (CNN+RNN) with a TimeDistributed layer. Our architecture combines Visual Geometry Group 19 (VGG19) for feature extraction with a Gated Recurrent Unit (GRU) for sequential analysis of real-time video data. While this design is well suited for temporal inputs, the TimeDistributed layer complicates gradient propagation and weakens the link between spatial and temporal information, reducing the effectiveness of standard visualization methods. To address this challenge, we adapt techniques such as saliency maps and Gradient-weighted Class Activation Mapping (Grad-CAM) to models that incorporate a temporal dimension. Although dedicated visualization methods for such architectures remain limited, our study highlights both the difficulties and the potential of applying tools originally designed for static images to recurrent convolutional networks handling video sequences. This approach extends classical interpretation strategies to temporal models and provides an intermediate solution until more specialized methods are developed.

\vspace{0.5cm}
 Gradient-weighted Class Activation Mapping
\end{abstract}

\newpage
\noindent\textbf{Highlights}
\begin{itemize}
    \item Adaptation of Grad-CAM and saliency maps to CNN+RNN with TimeDistributed layer.
    \item Identification of limits in classical visualization for sequential models.
    \item Contour-based visualization method to highlight model attention areas.
    \item Visualizations supporting optimization of GRU size, dropout rate, and MLP depth.
\end{itemize}

\begin{figure}[H]
    \centering
    \includegraphics[width=\textwidth]{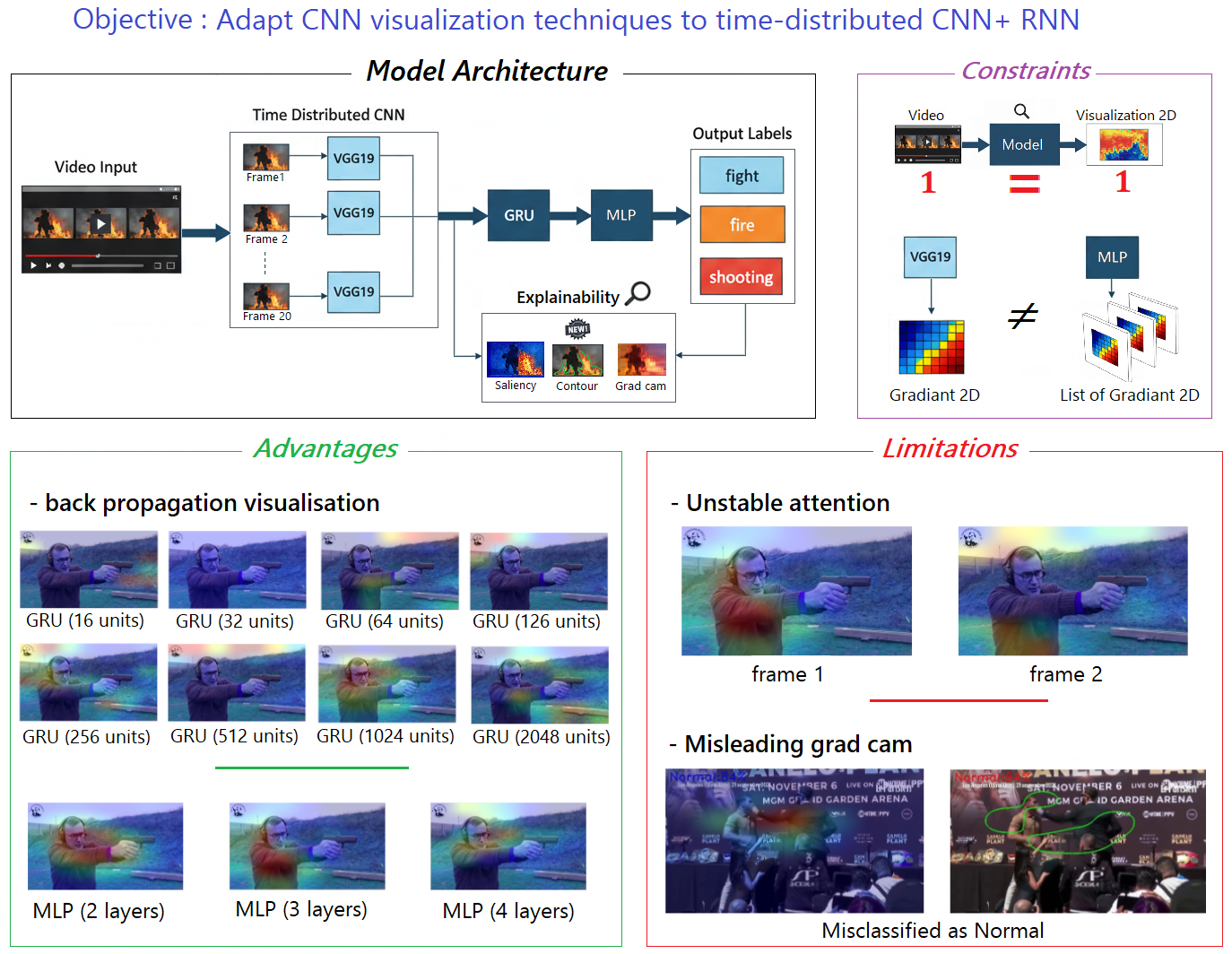}
    \caption{Graphical abstract}
    \label{fig:graphical_abstract}
\end{figure}

\newpage
\section{Introduction}
\label{sec:introduction}
Deep neural networks play a central role in many fields, particularly in solving complex problems such as real-time anomaly detection in video. However, their effectiveness comes with significant opacity: they are often seen as "black boxes", unable to explain or justify their decisions. This opacity presents critical challenges, especially in sensitive areas like security, where anomalies such as fights, gunshots, or accidents need to be detected quickly and reliably. The lack of transparency (explainability) raises ethical concerns about responsibility for automated decisions, as well as legal issues, particularly with the implementation of the General Data Protection Regulation. This regulatory framework, and in particular Article 22~\cite{rgpdArticle22}, requires significant decisions not to be based solely on automated processing, which makes the explainability of such systems essential.
In this context, the explainability of deep learning models that process video sequences through a TimeDistributed architecture remains a major challenge.  Although visualization techniques are well established for Convolutional Neural Networks (CNN) applied to single images, their adaptation to CNN–RNN\footnote{Recurrent Neural Networks} models with a temporal dimension is still underexplored. The temporal dimension of video data challenges traditional visualization methods, raising the question of how to make the decisions of Time-Distributed models for anomaly detection visible and interpretable.
To address this issue, we explore visualization techniques, such as saliency maps and Gradient-weighted Class Activation Mapping (Grad-CAM), applied to a model that combines CNNs for spatial feature extraction with Recurrent Neural Networks (RNNs) for temporal analysis within a TimeDistributed architecture. Our model, developed using the Keras library\footnote{see~\href{https://keras.io}{keras.io}}, uses convolutional layers based on the Visual Geometry Group 19 (VGG19) architecture for visual feature extraction and Gated Recurrent Unit (GRU) layers for sequential analysis, allowing real-time video processing.
The TimeDistributed architecture enables each frame of a video sequence to be analyzed individually while preserving temporal continuity, making it particularly well suited for real-time analysis of complex videos~\cite{10479307}. 
However, this structure leads to specific challenges. It complicates the propagation of the gradient because each element in the sequence is processed by the same submodel independently, making it more difficult to associate gradients with particular temporal elements. Therefore,we adapt visualization techniques such as saliency maps and Grad-CAM to take into account the temporal dimension in models that process video.

This study presents two major contributions. First, we analyze the limitations of current visualization techniques and highlight the specific constraints imposed by the Time-Distributed architecture, particularly with respect to gradient propagation and the generation of interpretable maps. Second, we propose the adaptation of visualization techniques originally developed for convolutional networks so that they can also be applied to models that incorporate temporal information. Finally, we show that explainability can also guide design choices in model development, for example by informing hyperparameter selection for systems intended to operate in real time.


\section{Related works}
\label{sec:relatedWork}
Surveillance videos, often devoid of audio, led us to focus our exploration on technologies applicable to images, which align with the type of data processed by the CNN component of our model. Moreover, the decisions made by the CNN can be visually represented, making them more intuitive and accessible, even to non-expert audiences. 
In contrast, the RNN relies on abstract temporal relationships, making the explanation of its decisions significantly more complex. 
In this context, the book by Christoph Molnar, "A Guide for Making Black-Box Models Explainable", published in 2019, serves as a key reference. 
It synthesizes and analyzes various explainability and visualization approaches used to interpret complex machine learning models~\cite{molnar2019}. 
This book marks a significant milestone in popularizing and structuring explainability techniques, highlighting their strengths, limitations, and areas of application, providing a valuable framework for guiding research in this domain.

On the one hand, some technologies, such as LIME (Local Interpretable Model-Agnostic Explanations), proposed on August 9, 2016, by Marco Tulio Ribeiro et al.~\cite{ribeiro2016should}, or SHAP (SHapley Additive exPlanations), introduced in November 2017 by Scott M. Lundberg and Su-In Lee~\cite{lundberg2017unified}, stand out due to their independence from the internal structure of the models. 
These methods work by analyzing the inputs/outputs of a model to assign local or global importance to input features. For instance, SHAP, based on the Shapley value theory~\cite{algaba2019handbook}, provides a coherent and unified explanation of each feature impact on the prediction, while LIME generates locally simplified models to interpret specific predictions. 
These technologies are particularly useful in environments where models are complex or proprietary, and their flexibility allows them to be applied to a variety of models, whether linear, ensemble-based, or neural networks. However, their implementation often requires the installation of specific libraries, along with pre- or post-processing steps to make the models compatible with these tools.

On the other hand, some technologies are specific to certain types of model, particularly CNNs, which primarily process structured data such as images or videos. For these models, visualization techniques are used that directly exploit their internal architecture. 

\begin{figure}[h!]
\centering
\begin{tikzpicture}[
    every node/.style={
        rectangle, draw, rounded corners,
        align=center, inner sep=5pt, font=\footnotesize
    },
    root/.style={text width=4.0cm, font=\bfseries}, 
    box/.style={text width=3.5cm}                    
]

\node[root] (root) {Explainability\\approaches};

\node[box, text width=4cm] (act) at (4,3) {
    \begin{minipage}{\linewidth}
    \centering
    \textcolor{blue}{Discriminative regions \\ (Activation Maps):} \\
    \rule{\linewidth}{0.4pt} \\[2pt]
    Grad-CAM, Grad-CAM++, LayerCAM, 3D extension
    \end{minipage}
};

\node[box, text width=4cm] (sal) at (-5,3) {
    \begin{minipage}{\linewidth}
    \centering
    \textcolor{blue}{Pixel importance \\ (Saliency Maps):} \\
    \rule{\linewidth}{0.4pt} \\[2pt]
    Vanilla, SmoothGrad, Guided Backprop, LRP, Integrated Gradients, DeepLIFT, Ply-CAM, Eigen-CAM, 3D extension
    \end{minipage}
};

\node[box, text width=4cm] (att) at (-5,-3){
    \begin{minipage}{\linewidth}
    \centering
    \textcolor{blue}{Model including \\ attention mechanism \\ (Attention Maps):} \\
    \rule{\linewidth}{0.4pt} \\[2pt]
    Attention Map, CDAM, Smooth CDAM, Integrated CDAM, 3D extension
    \end{minipage}
};

\node[box, text width=4cm] (oth) at (4,-3) {
    \begin{minipage}{\linewidth}
    \centering
    \textcolor{blue}{Other techniques:} \\
    \rule{\linewidth}{0.4pt} \\[2pt]
    RISE, Occlusion sensitivity, DeconvNets, Toolkits
    \end{minipage}
};

\draw (root) -- (act);
\draw (root) -- (sal);
\draw (root) -- (att);
\draw (root) -- (oth);

\end{tikzpicture}
\caption{Taxonomy of explainability methods for CNN-based models.}
\label{fig:taxonomy}
\end{figure}

Figure~\ref{fig:taxonomy} summarizes the main families of explainability approaches (activation, saliency, attention, and other techniques). The following subsections describe each of them, with their representative methods, variants, and extensions.

\newpage\subsection{Activation Maps and Their Variants}
\begin{itemize}
    \item Gradient-weighted Class Activation Mapping (Grad-CAM): Generates heatmaps by weighting the activations of the last layers with gradients to highlight the regions important for classification~\cite{selvaraju2017grad, Fayyaz2025}.

    \item Grad-CAM++: An improvement on Grad-CAM that considers higher-order interactions for more precise maps, particularly in cases where multiple objects or classes are present in the image~\cite{aditya1710grad}.

    \item LayerCAM: Another method that generates activation maps using activations from shallow layers of the network to provide more accurate localization information~\cite{jiang2021layercam}.

\end{itemize}

\subsection{Saliency Maps and Their Improvements}
Saliency maps identify the regions of an image that have the greatest impact on the model's decision~\cite{simonyan2014deep}. Several techniques have been developed to improve their quality:
\begin{itemize}

    \item SmoothGrad: Reduces noise by averaging gradients calculated over slightly perturbed inputs~\cite{smilkov2017smoothgrad}.

    \item Guided Backpropagation: Combines activation gradients with forward propagation information for sharper maps~\cite{springenberg2014striving}.

    \item Layer-wise Relevance Propagation (LRP): Assigns relevance to each neuron by following the decision flow through the layers of the network to form a saliency map~\cite{bach2015pixel}.

    \item Integrated Gradients: Creates saliency maps by calculating the importance of pixels by integrating gradients between a reference input and the actual input~\cite{sundararajan2017axiomatic}.

    \item Deep Learning Important FeaTures (DeepLIFT): Compares a neuron's activations with a reference activation to generate more understandable saliency maps~\cite{shrikumar2017learning}.

    \item Poly-CAM: This method generates saliency maps by recursively merging activation maps from early layers with those from deeper layers for more detailed visualizations~\cite{englebert2024poly}.

    \item Eigen-CAM: Uses the decomposition of activations into principal components to generate saliency maps while integrating gradients to refine discrimination between classes~\cite{muhammad2020eigen}.

    \item Fast-CAM: A fast method to generate saliency maps, using Grad-CAM or Grad-CAM++ for better computational efficiency~\cite{mundhenk2019efficient}.

\end{itemize}

\subsection{Attention Maps and Their Improvements}
\begin{itemize}
    \item Attention Maps (AM): These maps visualize the input areas that a model pays the most attention to, typically used in models with an attention mechanism such as Transformers. They help identify influential regions in the model's decisions by exploiting self-attention weights calculated in deep layers~\cite{brocki2023class}.

    \item Class-Discriminative Attention Maps (CDAM): An extension of classic attention maps, CDAM adjusts the attention scores to make them specific to a target class. This is achieved by weighting attention scores by the relevance of the "tokens" associated with a class, using gradients. CDAM is particularly effective for providing compact and semantic explanations by targeting specific concepts~\cite{brocki2023class}.

    \item Smooth CDAM: A variant of CDAM that adds Gaussian noise to the input "tokens" to generate averaged explanations, thus improving the robustness of attention maps~\cite{brocki2023class}.

    \item Integrated CDAM: Another improvement of CDAM that interpolates between a reference input and the actual input to produce integrated explanations, inspired by integrated gradients. This improves the stability of explanations, particularly for latent or complex concepts~\cite{brocki2023class}.
\end{itemize}

\subsection{Other Techniques}
\begin{itemize}
   \item RISE (Randomized Input Sampling for Explanation): A method that generates importance maps by applying numerous randomized binary masks to the input and aggregating the model responses. RISE can be extended to video (RISE-Video) to produce spatiotemporal importance maps by masking spatial regions and/or entire frames~\cite{petsiuk2018riserandomizedinputsampling}. Importantly, RISE is versatile and can be applied to any of the aforementioned visualization methods, including 3D networks and attention maps, providing an additional tool to interpret complex models.

    \item Extensions to 3D networks: Garcia-Torres and Meinich-Bache (2024) adapt Grad-CAM and Saliency methods to 3D CNNs~\cite{fernandez2024interpretability}. Originally designed for volumetric or spatiotemporal data such as videos or 3D medical images, these methods extend classical attention maps by incorporating both spatial and temporal dimensions, allowing the localization of important regions in complex data such as magnetic resonance image (MRI) scans or video sequences~\cite{lozupone2024axial}. In particular, these improvements can also be applied to attention maps in transformer-based architectures.

    \item Occlusion Sensitivity: A technique in which portions of the input are masked to observe how this affects the prediction of the models, providing information about important regions~\cite{uchiyama2023adaptiveocclusionsensitivityanalysis}.

    \item DeconvNets: Use a deconvolution process to reverse the activation process of a network, allowing more intuitive visualizations of intermediate layer activations~\cite{zeiler2013visualizingunderstandingconvolutionalnetworks}.
\end{itemize}
      
\noindent Many libraries exist to help users visualize the features learned by neural network models, particularly in the fields of computer vision and medical learning. In 2017, Kotikalapud Raghavendra proposed keras-vis~\cite{raghakotkerasvis}, a public library that allows users to visualize the convolution filters of each layer, their evolution during training, and the activation maps. Later, in 2020, Philippe Remy developed keract~\cite{Keract}, another library that performs similar operations. 
In 2020, Gotkowski et al. proposed a library that allows users to visualize both 2D and 3D attention maps in deep learning models for volumetric or spatiotemporal data, particularly in medical applications~\cite{gotkowski2020m3d}. Furthermore, the Keras library, developed by François Chollet in 2015~\cite{chollet2015keras}, now includes some of these visualization techniques. 
Another notable library is M3d-CAM, which generates 3D attention maps for medical deep learning applications, facilitating the visualization of relevant areas in 3D images such as MRIs or CT scans~\cite{gotkowski2021m3d}.

In the field of explainable Artificial Intelligence (XAI) for computer vision, several toolkits and review studies have been proposed to enhance model transparency. Abdullah et al. (2024) introduced VisionCam, a comprehensive XAI toolkit designed to interpret image-based deep learning models~\cite{abdullah2024visioncam}. Furthermore, two recent studies provide in-depth analyses of explainable models: Mahmoudi et al. (2023) offer a comparative study of deep learning models applied to real-time action recognition~\cite{mahmoudi2023review}, while C. Roy et al. (2023) present a probabilistic approach to explainable video activity recognition~\cite{roy2023explainable}.


\section{Our Approach for Real-time Anomaly Detection}
\label{sec:approach}
The model used in this study has already been described in a previous publication~\cite{10479307}. Although more recent architectures, such as Transformers, or autoencoders, can achieve strong performance in video analysis, their high computational cost and inference latency make them unsuitable for our real-time anomaly detection context with limited resources. For this reason, we have opted for a CNN + RNN architecture developed using the Keras library, specifically designed for anomaly detection in video sequences. The architecture consists of a convolutional component (VGG19) for feature extraction and a recurrent component (GRU with 1024 neurons) for temporal relationship analysis. A multilayer perceptron (MLP) composed of three dense layers (with 1024, 512, and 128 neurons, respectively) is employed, where each dense layer incorporates a dropout rate of 50\% and L2 regularization with a coefficient of 0.01. 
The VGG19 model is applied to each frame via a TimeDistributed layer to preserve temporal ordering. 
The choice of VGG19 as the convolutional backbone was established through an empirical comparison against several ImageNet-pretrained architectures, including Inception, Inception-ResNet, ResNet, VGG-Place325, Xception, and EfficientNet, reported in our earlier work~\cite{poirier2024realtimeanomalydetectionvideo}. Among the backbones tested, VGG19 consistently achieved the best F1-score on our dataset and was therefore retained for this study; we refer the reader to~\cite{poirier2024realtimeanomalydetectionvideo} for the complete comparative results and the corresponding methodology. 
The model is compiled using the Stochastic Gradient Descent (SGD) optimizer with a learning rate of 0.01 and trained with a batch size of 8 for 200 epochs. 
The dataset distribution is presented in Table~\ref{tab:dataset}, showing the number of videos and the total duration for each class.

For anomalous classes, training and validation videos were manually segmented to isolate specific anomalous events. In contrast, unsegmented videos containing no anomalies were added to establish a dedicated normal class. For these training and validation sets, input frames are resized to $112 \times 112$ pixels and grouped into 20-frame sequences. To stream these sequences efficiently during training, we employed a video generator\footnote{cf.~\href{https://github.com/Fab16BSB/keras-video-generators}{Source code for the video generator employed in this study.}} that feeds data in batches. Furthermore, a data augmentation pipeline including random zooms, horizontal flips, and brightness adjustments was applied to the training data to increase variability, enabling the model to handle diverse lighting conditions (day and night) as well as actions coming from the left or right.
In contrast, a held-out test set was kept completely untouched to evaluate the final performance of the model. Consisting of roughly thirty videos (ten per class), this test set was deliberately left unsegmented and unfiltered to reflect realistic, real-world deployment conditions rather than the curated clips used for training and validation.

\begin{table}[h]
\centering
\caption{Distribution of the dataset (in italic, dedicated normal class)}
\vspace{0.25cm}
\resizebox{\textwidth}{!}{%
\begin{tabular}{c|c|c|c|c}
    \hline
    Classes   & Train Videos (nb) & Train Duration & Validation Videos (nb) & Validation Duration \\ \hline
    Fight    & 587          & 0h50           & 391               & 0h31 \\
    Fire     & 237          & 3h40           & 61                & 0h46 \\
    Shooting & 247          & 0h17           & 64                & 0h08 \\ 
    \emph{Normal} & \emph{760}  & \emph{5h44}  & \emph{445}      & \emph{3h21} \\
\end{tabular}%
}
\label{tab:dataset}
\end{table}

The dataset exhibits a clear class imbalance, both among anomaly classes and relative to the Normal class. This is an inherent characteristic of the task rather than a data collection artifact: everyday, non-anomalous activity is comparatively easy to source and naturally serves as the reference for the Normal class, whereas anomalies are by definition rare events, and some anomaly types occur less frequently than others in real-world footage. Fight is comparatively well represented, whereas Fire and Shooting are markedly less frequent, reflecting the relative rarity of these events in the sources used to build the dataset.

\begin{figure}[H] 
 \begin{center} 
    \includegraphics[width=\textwidth]{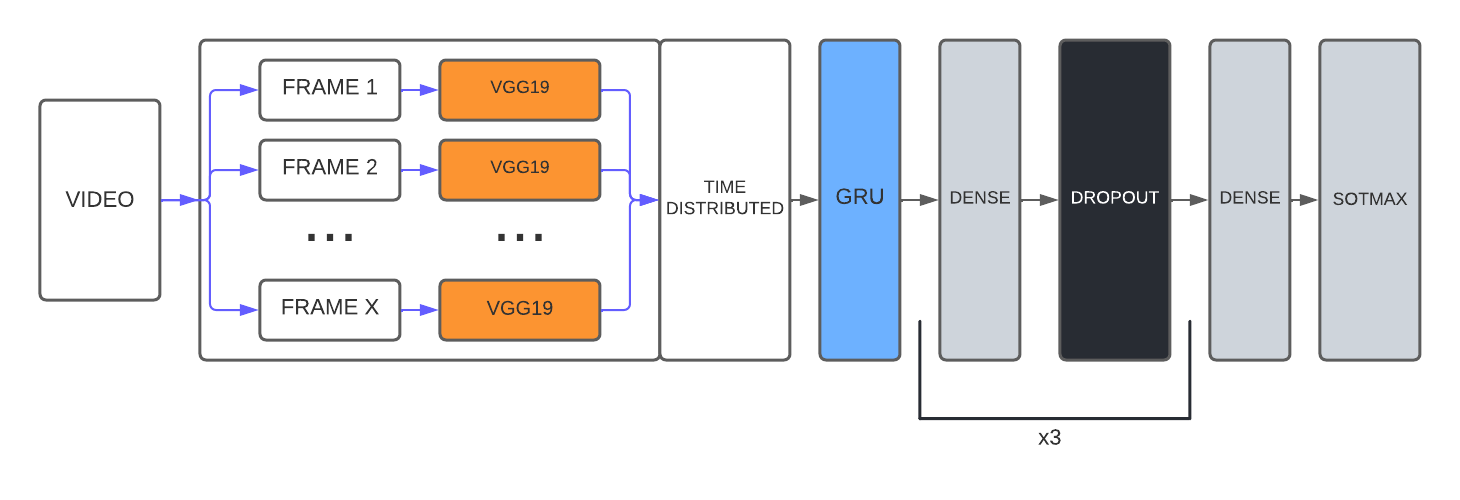} 
    \caption{Model Architecture} 
    \label{Model} 
 \end{center} 
\end{figure}

\noindent On the one hand, two-dimensional (2D) convolutions are commonly used in image processing. 
They apply a filter to an image to extract local features such as edges, textures, or patterns. Each filter is applied to a spatial dimension of the image (width and height), without considering the temporal dimension. Thus, a 2D convolution treats an image as a single input, without temporal relation between multiple images in a sequence. \\

\noindent Three-dimensional (3D) convolution, on the other hand, is specifically designed for processing volumetric data or videos, where the input includes both spatial dimensions (width and height) and a temporal dimension (depth or the number of images in a sequence). 
Unlike 2D convolutions, which focus on processing each image independently, 3D convolutions apply filters simultaneously on both the spatial and temporal dimensions. 
This allows the model to capture features not only within each image but also across relationships between consecutive images. 
Thanks to this capability, 3D convolutions are particularly suited for identifying spatiotemporal patterns, tracking the evolution of visual information over time, which is crucial for understanding movements, transitions, or dynamic changes in a sequence. \\

\noindent The Time Distributed layer allows the same layer, such as VGG19, to be applied independently to each image of a video sequence while maintaining the temporal integrity of the data. In other words, it "distributes" the processing of each image independently while respecting the temporal order of the data. The difference in the context with 2D or 3D convolutions is that, in the case of Time Distributed, each image in the sequence is treated identically, but the model retains the temporal information through the sequence structure, without applying convolution on the temporal dimension itself. This allows for independent extraction of visual features while preserving the order of the images for further processing by recurrent layers like GRU. \\

\begin{figure}[H] 
    \begin{center} 
        \begin{minipage}[b]{0.32\textwidth}                
            \includegraphics[width=\textwidth]{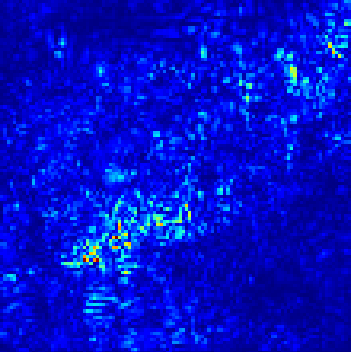} 
            \caption{Saliency maps} 
            \label{saillance} 
        \end{minipage} 
        \hfill 
        \begin{minipage}[b]{0.45\textwidth} 
            \includegraphics[width=\textwidth]{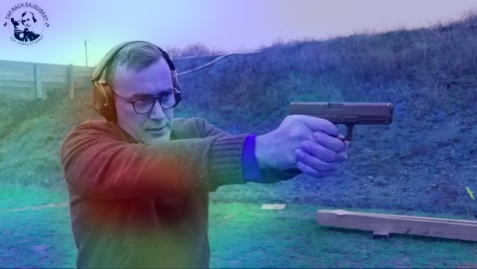} 
            \caption{Activation maps \\ Source “tap rack gajoubert“} 
            \label{activation} 
        \end{minipage} 
    \end{center} 
\end{figure}

\noindent To better understand the internal workings of our model, we opted for visualization techniques suited to CNNs, such as saliency maps (cf. Figure~\ref{saillance}), activation maps (cf. Figure~\ref{activation}) and filter visualizations. These methods, based on gradient propagation, allow for a direct exploration of the activations of network. 
Saliency maps identify the most influential pixels for a prediction, while activation maps reveal the specific regions of an image that contribute to the final decisions. 
By avoiding external dependencies, we ensure consistency and reproducibility in our analyzes.

\noindent Unfortunately, due to our specific architecture, we found that visualization libraries designed for classic CNNs were not suitable for our model. 
This limitation is directly related to the structure and peculiarities of our approach. \\

\begin{figure}[H] 
    \begin{center} 
        \includegraphics[width=\textwidth]{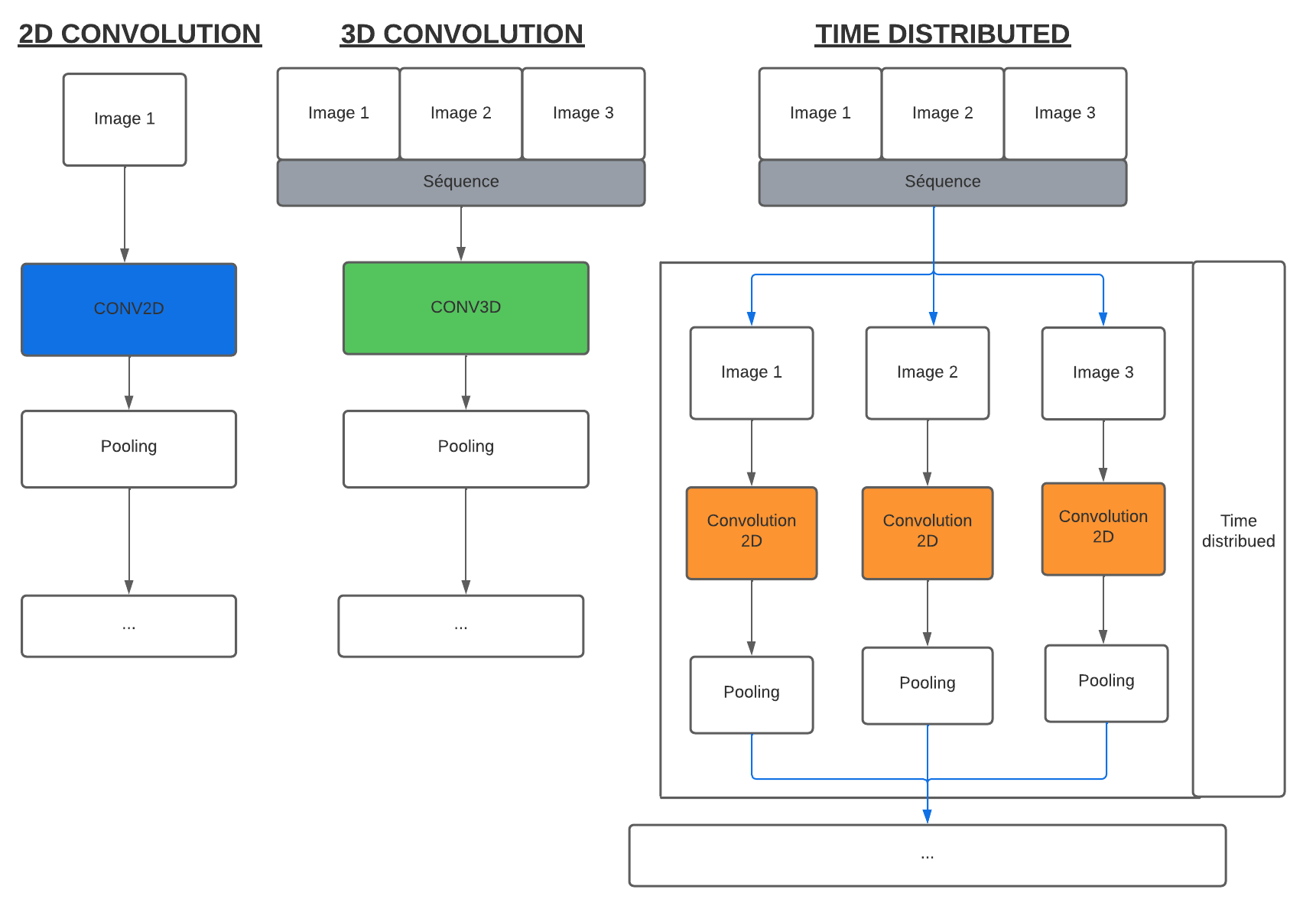} 
        \caption{Structure of 2D convolution / 3D convolution / Time distributed convolution } 
        \label{convolution comparison} 
    \end{center} 
\end{figure}

\noindent The first challenge lies in the organization of our model. Unlike a standard network using 2D or 3D convolutions, where the layers are directly connected and the outputs of one layer serve as inputs to the next, our architecture encapsulates the convolutions within a \textit{TimeDistributed} layer (see Figure~\ref{convolution comparison}). This structural choice allows us to apply VGG19 (our CNN) independently to each frame of the input video, thereby preserving the temporal organization of the data.  

However, this organization has a direct impact on gradient computation, which is crucial for model interpretation. Keep in mind that a gradient quantifies how a small variation in an input pixel influences the output of the model. Pixels with higher gradients are therefore more decisive for prediction, and the ability to compute them reliably is essential for generating meaningful visualizations. 

In a sequential setup, the gradients would be expected to reflect the influence of each convolutional layer on the prediction. Yet, because the \textit{TimeDistributed} wrapper processes each frame separately, backpropagation behaves differently. Instead of propagating gradients layer by layer over time, it computes them independently for each frame and then aggregates the results into a single value (cf. Equation~\ref{eq:gradient}):
\begin{equation}
\label{eq:gradient}
 \frac{\partial L}{\partial W} = \sum_{t=1}^T \frac{\partial L_t}{\partial W}   
\end{equation}

where $L$ represents the total loss (model error), $W$ the internal parameters of the network, and $L_t$ the loss at the time step $t$. \\

This aggregation has an important drawback: it obscures the role of individual convolutional layers inside the \textit{TimeDistributed} block. The resulting gradient is merely a global sum, making it difficult to isolate the specific contribution of each encapsulated layer to the prediction. As a consequence, understanding and visualizing the internal decision process of the network becomes considerably more challenging.  

\noindent The second challenge comes from the addition of the temporal dimension. Unlike models based on 2D or 3D data, which generate a single output visualization for a single input (such as an image or volumetric object), our model processes video sequences, which involve multiple input images for a single output sequence. 
This paradigm shift requires rethinking how visualizations are generated and interpreted, as each input image contributes to the overall prediction of the sequence.

\noindent These limitations directly affect the visualization methods. To generate saliency maps, we had to compute gradients for each image in the sequence, resulting in a series of gradients corresponding to the length of the sequence. 
These gradients are then individually displayed, generating a distinct saliency map for each image. For activation maps, the solution we implemented relies on using the output from the TimeDistributed layer. 
This layer applies identical processing to each image in the sequence, generating a list of outputs, each corresponding to an individual image. 
The calculated gradients have the same dimensions as these outputs, which allows them to be associated image by image and projected onto the corresponding image. 
However, with this architecture, it is only possible to generate activation maps for the output layer of the submodel, which limits the analysis of the internal layers. Visualizing this single output provides a summarized view of the temporal activations used for anomaly detection but does not allow a direct examination of the intermediate representations learned by the convolutional layers. This limitation mainly arises from the use of the TimeDistributed layer: to compute the gradient of the loss with respect to the intermediate CNN activations, the model must propagate the gradient through the GRU and the TimeDistributed layer. However, each CNN layer produces only one vector per image, whereas the GRU expects a full sequence of gradients. This mismatch in dimensions prevents proper propagation of the gradient to certain convolutional layers.

\section{Experimentation}
\label{sec:experimentation}
\begin{figure}[H]
\begin{center}
 \begin{minipage}[b]{0.4\textwidth}
    \centering
    \includegraphics[width=\textwidth]{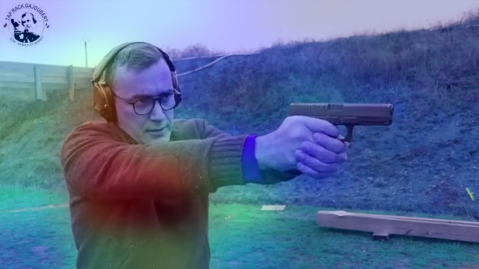}
    \small{(a)}
 \end{minipage}
 \hfill
 \begin{minipage}[b]{0.4\textwidth}
    \centering
    \includegraphics[width=\textwidth]{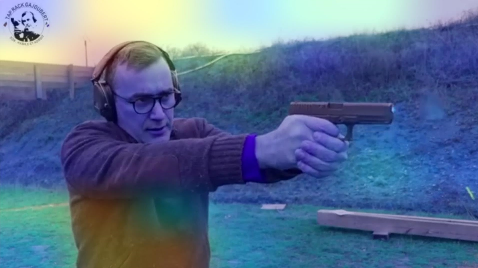}
    \small{(b)}
 \end{minipage}
 
 \vspace{0.5cm} 

 \begin{minipage}[b]{0.4\textwidth}
    \centering
    \includegraphics[width=\textwidth]{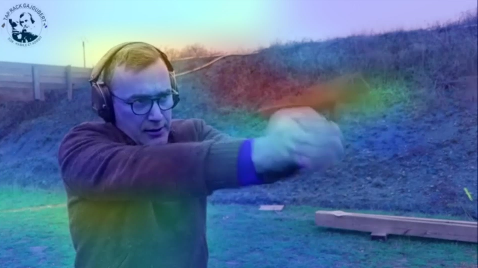}
    \small{(c)}
 \end{minipage}
 \hfill
 \begin{minipage}[b]{0.4\textwidth}
    \centering
    \includegraphics[width=\textwidth]{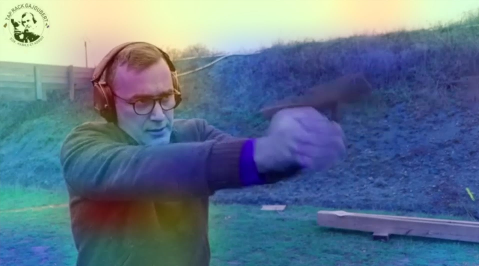}
    \small{(d)}
 \end{minipage}
\end{center}
\caption{Class activation maps (Grad-CAM) for a shooting video \\ Source: ``tap rack gajoubert``}
\label{activation_class}
\end{figure}

In this section, we will present our results, as well as the advantages and disadvantages of each approach explained previously (see Section~\ref{sec:relatedWork}). 
All of our visualizations were obtained from a proprietary dataset\footnote{The data sets used are confidential and cannot be shared publicly.}. 
Upon examination of the activation maps shown in Figure~\ref{activation_class}, which represent a sequence of images from a video that illustrates a gunshot, we can observe that our model does not consistently focus on the same areas of an image, even when the images are successive. 
This is due to the operation of the convolutional layers, which process each image independently, without explicitly linking important areas of one image to those of others in the sequence.

\begin{figure}[H]
\begin{center}
    \includegraphics[width=0.45\textwidth]{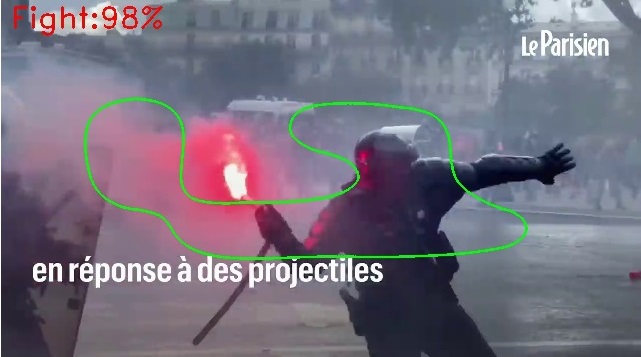} \hfill
    \includegraphics[width=0.45\textwidth]{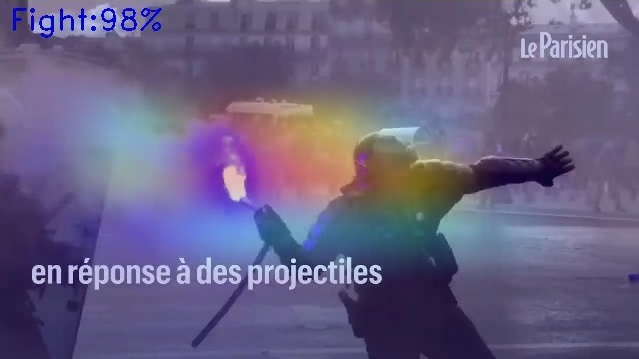} \\
    \small{(a)} \hspace{0.42\textwidth} \small{(b)}
    
    \vspace{0.5cm} 
    
    \includegraphics[width=0.45\textwidth]{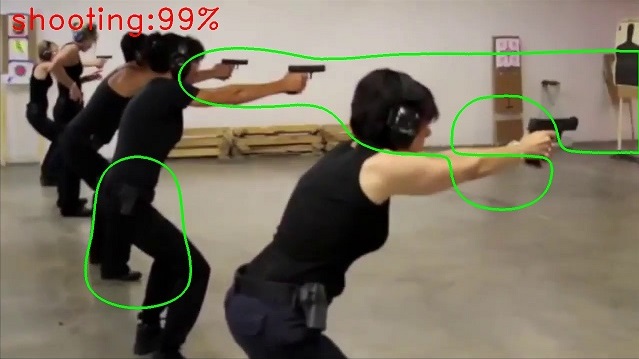} \hfill
    \includegraphics[width=0.45\textwidth]{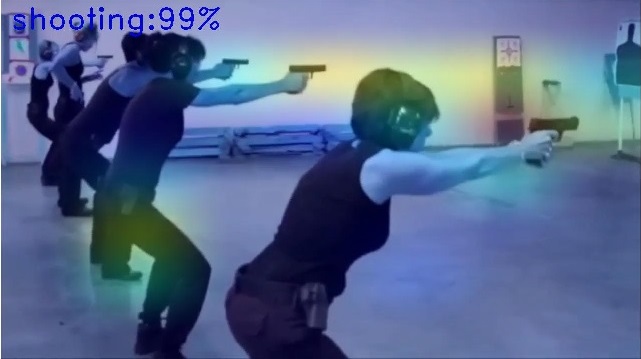} \\
    \small{(c)} \hspace{0.42\textwidth} \small{(d)}
\end{center}
\caption{Example of visualization for the classes: fight and shooting. \\
Source: ``Le Parisien newspaper``}
\label{border_visualisation}
\end{figure}

\noindent Analyzing activation maps (such as Grad-CAM) is a common technique to interpret decisions from deep learning models. However, we have found that these visualizations can be ambiguous or even misleading. In some cases, they can underrepresent crucial elements of the scene or give a false impression of confidence when the classification is actually incorrect (false negative). To overcome these limitations and enable for more detailed analysis, we developed a new visualization approach. This method involves extracting the contours corresponding to the different activation zones using the "Open Source Computer Vision" (OpenCV) library\footnote{see \href{https://opencv.org/}{opencv.org}}. 
Figure~\ref{border_visualisation} illustrates the basic principle of our method for correctly classified examples. Images \ref{border_visualisation}.(a) and \ref{border_visualisation}.(c) show the classic activation maps for the fight and shooting classes. Images \ref{border_visualisation}.(b) and \ref{border_visualisation}.(d) present our contour visualization applied to these same scenes, clearly delineating the areas on which the model focuses its attention.

\begin{figure}[H]
\begin{center}
    \includegraphics[width=0.45\textwidth]{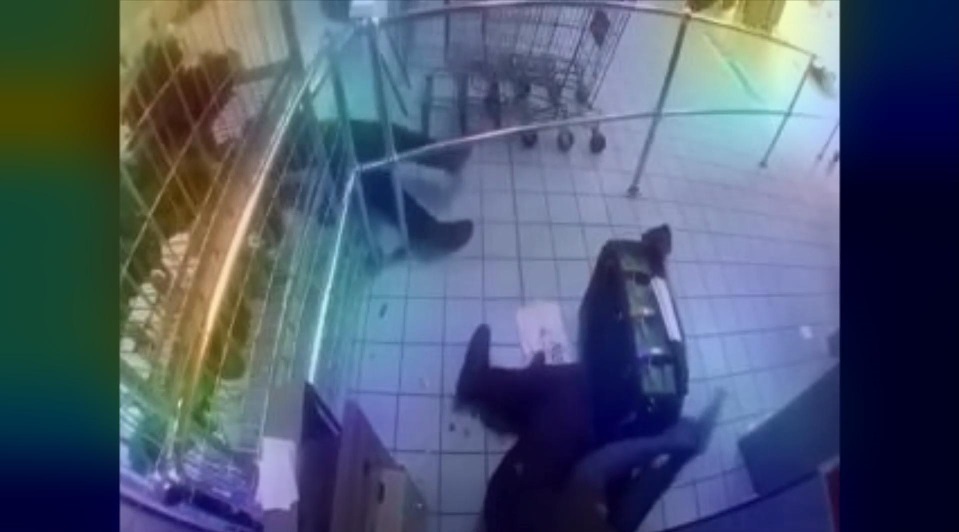} \hfill
    \includegraphics[width=0.45\textwidth]{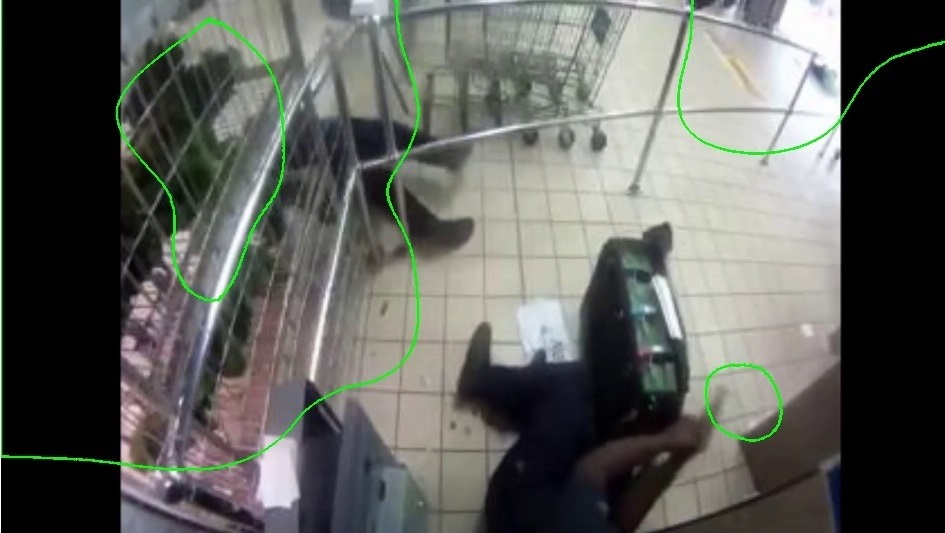} \\
    \small{(a)} \hspace{0.50\textwidth} \small{(b)}
    
    \vspace{0.5cm} 
    
    \includegraphics[width=0.45\textwidth]{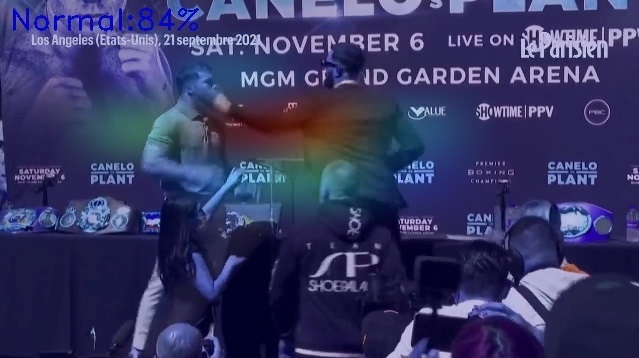} \hfill
    \includegraphics[width=0.45\textwidth]{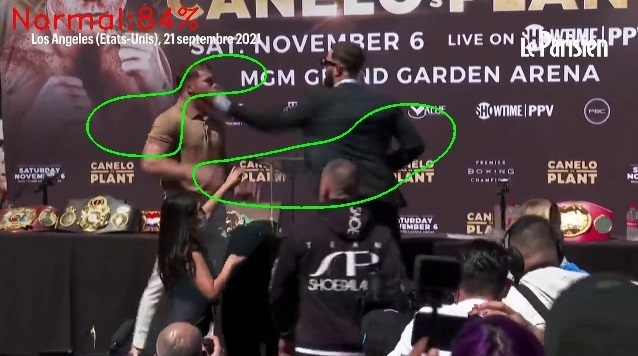} \\
    \small{(c)} \hspace{0.50\textwidth} \small{(d)}
\end{center}
\caption{Activation map and contour visualization applied to images extracted from shooting and fight videos. \\
Source: UCF Crime + ``Le Parisien newspaper``}
\label{border_advantage}
\end{figure}

\noindent  The main advantage of our technique lies in its ability to reveal the model's interpretative weaknesses where traditional activation maps fail. Figure~\ref{border_advantage} highlights the cases where our method exposes these problematic interpretations. For example,  Image~\ref{border_advantage}.(a) shows the activation map for a shooting scene. Although the firearm is an essential element, it generates only a low activation, which is hardly perceptible on the map. However, our contour visualization in the image~\ref{border_advantage}.(b) successfully isolates and clearly delineates this object, confirming that the model did notice it, albeit weakly. The case illustrated by the images~\ref{border_advantage}.(c) and \ref{border_advantage}.(d) is even more revealing of a potential interpretation error. The activation map in the image~\ref{border_advantage}.(c) appears to correctly focus on the interaction between two individuals in a fight. Visually, one might conclude that the model has understood the scene correctly. However, the associated prediction was incorrect (a false negative, classified as "Normal"). The contour analysis in the image~\ref{border_advantage}.(d) exposes the reason for this failure: the model completely ignores the main action (the thrown punch) and instead focuses on the general silhouettes of people. 
Therefore, the alone activation map was misleading, while our contour visualization allowed us to precisely diagnose the model flaw.

\begin{figure}[H]
    \centering
    \includegraphics[width=13cm]{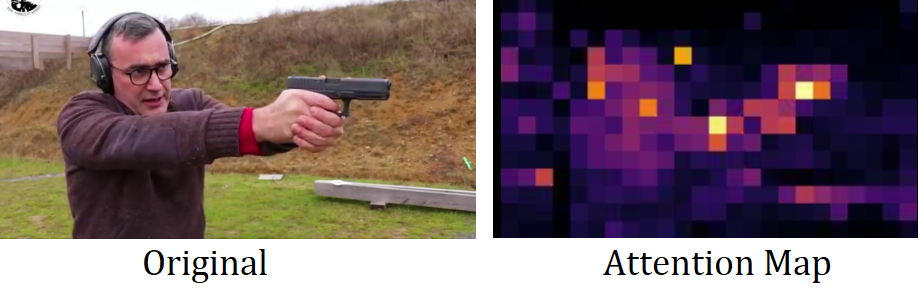}
    \caption{Attention map perform with self distillation with no label (DINO)~\cite{caron2021emerging}}
    \label{attention_map}
\end{figure}

\noindent Despite its advantages, our approach has some limitations. Edge detection can be inaccurate, especially when activation areas of different intensities are nested. Furthermore, this representation does not directly quantify the activation intensity, making it more qualitative than quantitative.
More generally, classical visualizations derived from CNNs, such as Grad-CAM, saliency maps, or activation contour visualizations, face several challenges. The lack of an attention mechanism in CNNs such as VGG19 leads to activated regions that vary from one image to another, even for successive, very similar images (see Figure~\ref{activation_class}). These variations complicate the interpretation of results and the comparison of activations across frames.
In contrast, modern techniques such as Vision Transformers incorporate an attention mechanism that allows them to continuously focus on the relevant regions of an image. The attention maps they produce are therefore more stable and coherent (cf. Figure~\ref{attention_map}), providing a smooth visualization of regions of interest, which is particularly useful for the analysis of video sequences. 
Another limitation lies in the computational cost of generating such visualizations. Producing grad-map or activation contours for each scene significantly slows down execution, making real-time processing impractical. As a result, these visualizations are better suited for post hoc analysis to verify predictions or for monitoring during model training rather than for deployment in real-time applications.

\noindent Nevertheless, these visualization tools (Grad-CAM, saliency map, edge visualization) proved crucial to analyze and optimize our architecture. By observing the effect of different layers (GRU, Dense, Dropout) on the features learned by the convolutional part, we were able to tune our hyperparameters more efficiently. 
These adjustments, validated by metrics such as accuracy and recall, confirm the consistency between visual observations and quantitative results. 
First, we analyzed the impact of the number of neurons in the GRU layer. 

\begin{figure}[H]
    \includegraphics[width=\textwidth]{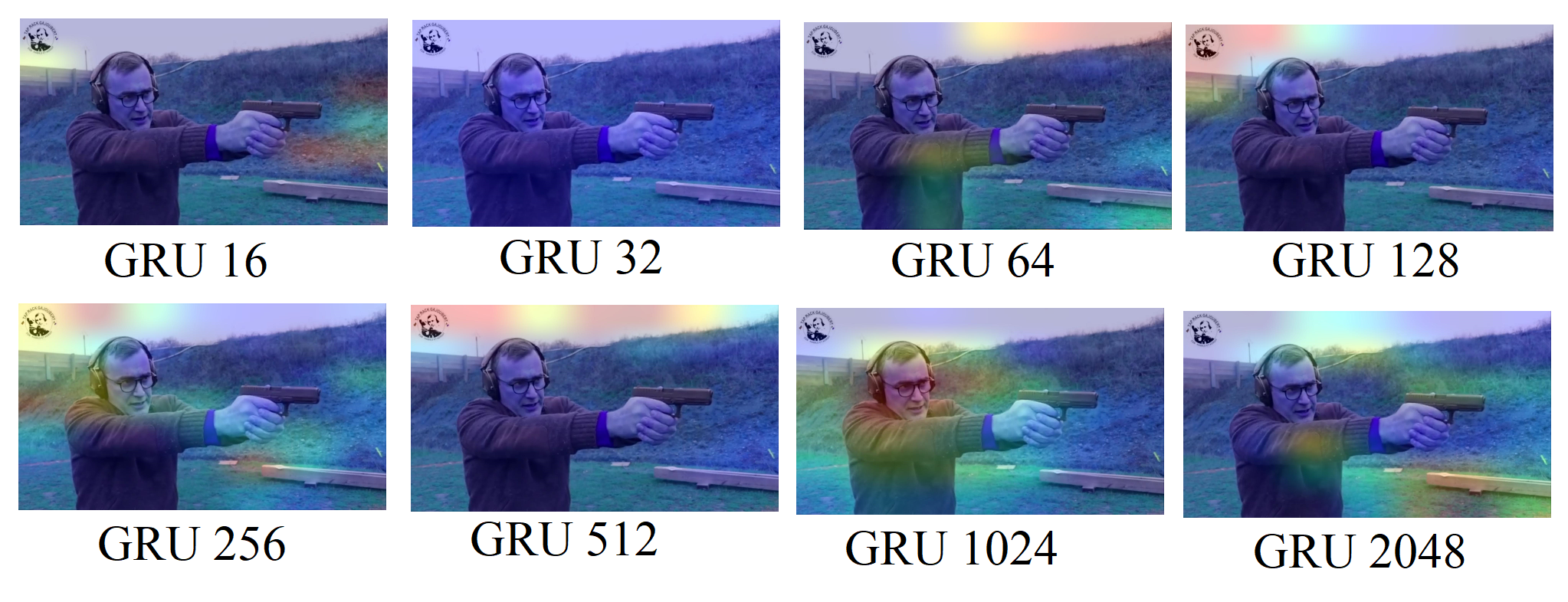}
    \caption{Activation map with variation of the number of neurons in the GRU Layer (50\% dropout). \\ Source “tap rack gajoubert“}
    \label{GRU_impact}
\end{figure}

\noindent The activation maps shown in Figure~\ref{GRU_impact} show a significant improvement in the concentration of activation areas for the 1024-neuron configuration. This is the only configuration where the activation area is clearly focused on the person in the video, which is crucial for anomaly detection. 
This visual observation is supported by the data in Table~\ref{tab:neurones_variation}, where this configuration achieves an optimal balance between precision (88.4\%) and recall (86.3\%). 

\begin{table}[H]
\centering
\begin{tabular}{|c|c|c|c|c|}
\hline
\textbf{Neurons} & \textbf{Accuracy} & \textbf{Precision} & \textbf{Recall} & \textbf{F1-score} \\ 
\textbf{(nb)} & \textbf{(\%)} & \textbf{(\%)} & \textbf{(\%)} & \textbf{(\%)} \\ 
\hline
8 & 83,3 & 19,6 & 84,2 & 31,7 \\ \hline
16 & 83,1 & 89,1 & 83,1 & 85.9 \\ \hline
32 & 82,9 & 91,7 & 82,9 & 87 \\ \hline
64 & 85,4 & 89,6 & 85,4 & 87.4 \\ \hline
128& 87,4 & 80,7 & 87,4 & 83.9 \\ \hline
256 & 85,3 & 81,1 & 85,3 & 83.1 \\ \hline
512 & 85,1 & 81,4 & 84,9 & 83.2 \\ \hline
\textbf{1024} & \textbf{86,3} & \textbf{88,4} & \textbf{86,3} & \textbf{87.3} \\ \hline
2048 & 83,4 & 84,0 & 83,4 & 83.6 \\ \hline
\end{tabular}
\caption{Impact of the number of neurons on performance with dropout rate set to 50\%.}
\label{tab:neurones_variation}
\end{table}

\noindent In contrast, lower or higher numbers of neurons lead to either a loss of information (dispersed activation) or overfitting (activation too focused on irrelevant areas). 

\begin{figure}[H]
\centering
    \includegraphics[width=13.5cm]{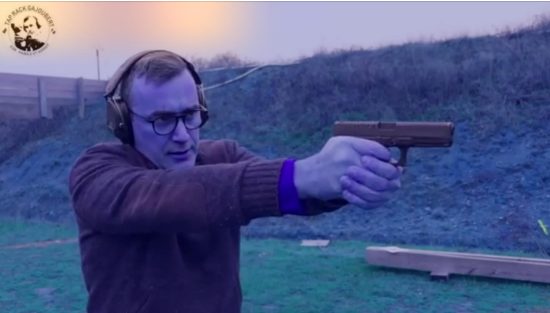}
    \caption{Activation map illustrating an overfit when using 1024 units without dropout} 
    \label{overfit}
\end{figure}

Then, we examine the effect of the dropout rate. Our findings, detailed in Table~\ref{tab:dropout_variation}, indicate that no dropout (0\%) leads to overfitting, as illustrated by Figure~\ref{overfit}. However, a dropout rate of 50\% provides an effective balance (cf. Figures~\ref{activation} and~\ref{activation_class}). This configuration maintains strong performance metrics concerning 89.6\% accuracy and 87.4\% recall while enhancing the model's robustness and its ability to generalize predictions to new data.

\begin{table}[H]
\centering
\begin{tabular}{|c|c|c|c|c|}
\hline
\textbf{Dropout} & \textbf{Accuracy} & \textbf{Precision} & \textbf{Recall (\%)} & \textbf{F1-score} \\ 
\textbf{(\%)} & \textbf{(\%)} & \textbf{(\%)} & \textbf{(\%)} & \textbf{(\%)} \\ 
\hline
0   & \textbf{92,8} & 83,0 & \textbf{92,8} & \textbf{87.6} \\ \hline
25  & 87,3 & 84,7 & 87,3 & 85.9 \\ \hline
50  & 86,3 & \textbf{88,4} & 86,3 & 87.3 \\ \hline
75  & 84,6 & 78,8 & 84,6 & 81.5\\ \hline
\end{tabular}
\caption{Performance as a function of the dropout rate for 1024 neurons.}
\label{tab:dropout_variation}
\end{table}

\noindent Notably, the alone quantitative metrics would have favored the opposite choice. Table~\ref{tab:dropout_variation} shows that the 0\% dropout configuration achieves the highest accuracy (92.8\%) and recall (92.8\%) of the four settings tested. 
Yet the activation map in Figure~\ref{overfit} shows a diffuse pattern, only loosely tied to the person and the weapon in the scene, consistent with overfitting rather than genuine discriminative learning. 
It was this discrepancy between a favorable metric and a visibly degenerate activation pattern that led us to retain the 50\% dropout configuration instead, despite its slightly lower raw scores. 
We report this as a concrete, if modest, illustration of visualization complementing quantitative metrics in a model selection decision, rather than a general validation of the approach.

\begin{figure}[H]
    \centering
    \includegraphics[width=\textwidth]{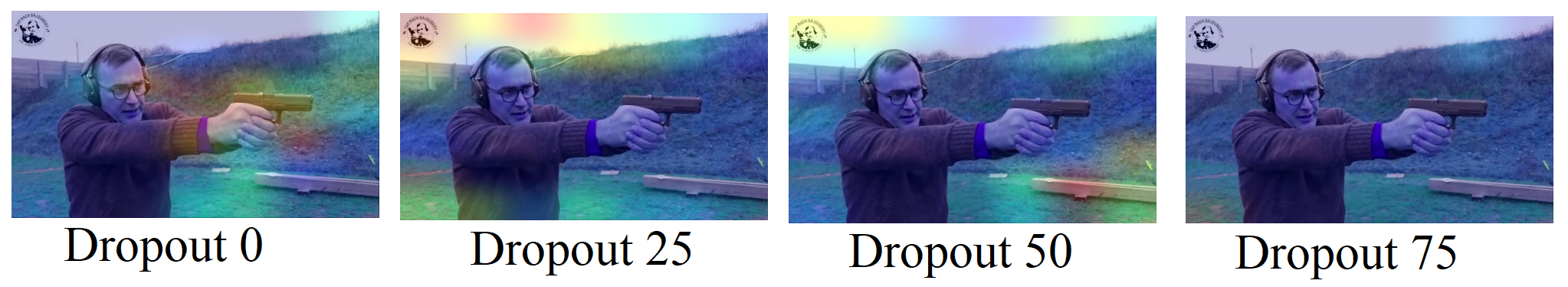}
    \caption{Activation map showing the variation of the dropout with an 64 neurons GRU. \\ Source “tap rack gajoubert“} 
    \label{dropout_impact}
\end{figure}

To help visualize how the dropout rate affects the activation areas of the model, we have included an activation map (see Figure~\ref{dropout_impact}). 
It is important to note that this visual is from a configuration with a 64-neuron GRU layer, not the 1024-neuron architecture we studied. 
This smaller configuration was chosen for illustration because it produces markedly more distinct activation patterns across dropout levels, making the qualitative effect of dropout easier to perceive visually. 
At 1024 neurons, the higher representational capacity of the layer tends to produce activation maps that look visually similar across dropout settings, which makes the effect harder to illustrate qualitatively. 
This image is therefore intended purely as a qualitative illustration of the dropout mechanism.  It should not be used to draw quantitative conclusions about the 1024-neuron configuration, which are instead supported by Table~\ref{tab:dropout_variation}.

\begin{figure}[H]
    \includegraphics[width=\textwidth]{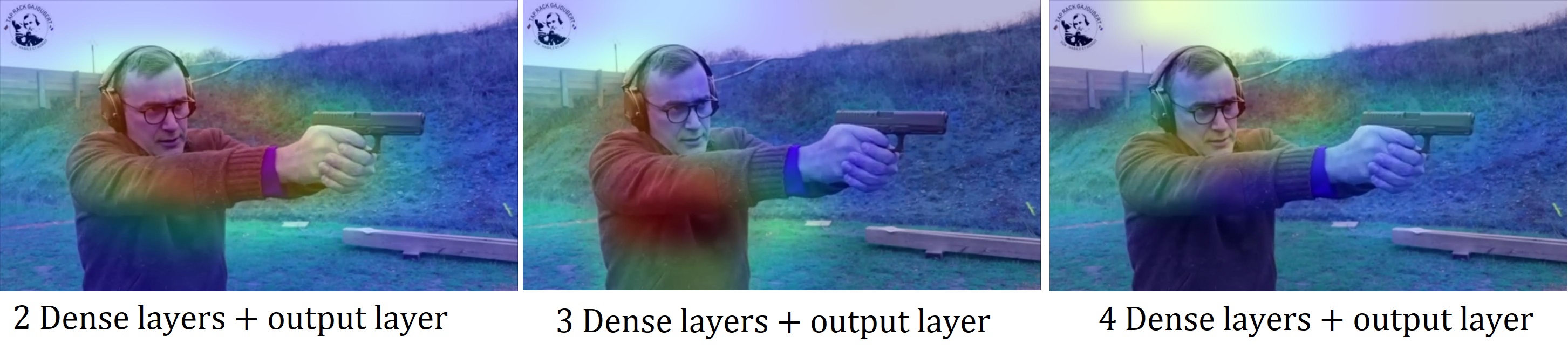}
    \caption{Activation map showing the variation of dense layer in the MLP using a GRU with 50\% dropout and 1024 neurons. \\ Source “tap rack gajoubert“}
    \label{dense_impact}
\end{figure}

\noindent Finally, we analyzed the effect of the number of output layers (cf. Figure~\ref{dense_impact}). 
The activation maps reveal significant differences between configurations. In the dense 2-layers configuration, the model focuses mainly on the face and posture of the person in the frame, indicating a limited focus on specific aspects. In contrast, with 3 dense layers, the activations are more evenly distributed across the person's overall posture, reflecting a more comprehensive understanding of the scene. 

\noindent This better concentration leads to an improvement in performance, as shown in Table~\ref{tab:output_layers}: the 3-layer configuration achieves an accuracy of 84.8\% and a recall of 92.2\%, outperforming the other configurations. 
The 4-layer configuration seems to introduce excessive complexity, reducing the coherence of activation, and resulting in a slight performance degradation. 

\begin{table}[H]
\centering
\begin{tabular}{|c|c|c|c|c|}
\hline
\textbf{Accuracy} & \textbf{Precision} & \textbf{Recall} & \textbf{F1-score} & \textbf{Blocks} \\ 
\textbf{(\%)} & \textbf{(\%)} & \textbf{(\%)} & \textbf{(\%)} & \textbf{ (Nb.)} \\\hline
89,8 & 84,2 & 89,8 & 86,9 & 2 \\ \hline
\textbf{92,2} & 84,8 & \textbf{92,2} & \textbf{88,3} & 3 \\ \hline
83,2 & \textbf{89,6} & 83,2 & 86,2 & 4 \\ \hline
\end{tabular}
\caption{Performance as a function of the number of blocks between GRU and output. Each block is composed of a dense layer and a dropout layer at 50\%.}
\label{tab:output_layers}
\end{table}

\begin{figure}[H]
        \centering
        \includegraphics[width=13cm]{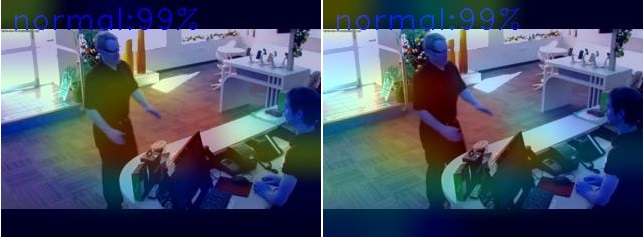}
        \caption{Activation Map for a normal video (images are sequential within the video)  \\}
        \label{activationNormal}
\end{figure}

\noindent This iterative approach, combining visualization and quantitative validation, allowed us to define an optimal configuration: 1024 neurons in the GRU, a dropout rate of 50\%, and 3 dense layers in the output. 
These adjustments ensure a robust high-performance model that is consistent with the visual observations. These visualizations also allow us to examine the features associated with the "normal" class. In a supervised learning context, this class represents the absence of anomalies, encompassing a wide range of actions such as working, walking, exercising, etc. It is interesting to note that movements in this class are often slow, unlike anomalies, which are characterized by sudden and rapid movements. 
Although a human would by default assign the normal class in the absence of anomaly indicators, our model does not follow this logic. 
To predict the normal class, it must detect the specific features that represent it. By visualizing examples of videos belonging to this class through activation and saliency maps, we found that our model mainly relies on the posture of the people on screen to predict this class. 
Observing the activation maps from successive frames (see Figure~\ref{activationNormal}), we can see that the activations target the hands of the person approaching the receptionist. 

\begin{figure}[H]
    \centering
    \begin{minipage}[b]{0.4\textwidth}
        \centering
        \includegraphics[width=5cm]{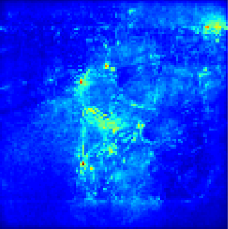}
        \large{(a)}
    \end{minipage}
    \hfill
    \begin{minipage}[b]{0.4\textwidth}
        \centering
        \includegraphics[width=5cm]{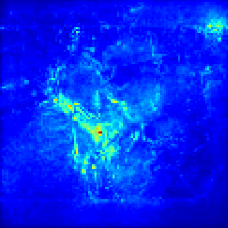}
        \large{(b)}
    \end{minipage}
    \caption{Saliency map for a normal video (images are sequential within the video).}
    \label{saliency}
\end{figure}

\noindent On the saliency map (see Figure~\ref{saliency}), the prominent pixels are mainly the outstretched arms in Figure~\ref{saliency}~(a), visible in the activation map, but we can also note the beginning of the shoulders and the lower body in Figure~\ref{saliency}~(b).

\section{Conclusions and Future Works}
\label{sec:conclusion}
In this article we have demonstrated how common image-based visualization techniques (activation maps, saliency maps, Grad-CAM and their variants) can be adapted to convolutional–recurrent architectures that use a TimeDistributed layer to process video sequences. 
The proposed adaptations provide interpretable frame-level visual cues that complement the model’s numerical outputs facilitating the optimization of architectural choices (for example, GRU size, dropout, MLP depth). At the same time, several important limitations must be acknowledged.
\vspace{0.25cm}

\noindent First, the visualization method implemented here is not yet suitable for continuous real-time deployment, as the generation and post-processing of activation maps introduce significant latency and computational overhead. 
Consequently, the visualization tools developed in this study are primarily intended for debug mode or offline analysis during model training and validation, where they help verify specific scenes and guide model development, rather than producing visual explanations for every frame in a live stream.
\vspace{0.25cm}

\noindent Second, due to the TimeDistributed+GRU architecture, it is currently possible to generate activation maps only for the output of the convolutional layer of the submodel. The visualization of this last convolutional layer gives a condensed frame-level view of the temporal activations used by the detector and delivers useful, interpretable cues about where the model focuses when predicting anomalies. 
However, it does not allow for direct access to intermediate convolutional representations. This restriction stems from a structural mismatch: computing gradients for intermediate CNN activations requires backpropagation through the GRU and TimeDistributed wrapper, creating a conflict between the frame-level vectors produced by the CNN and the sequence-level gradients required by the GRU. 
The resulting dimensionality mismatch prevents reliable propagation of the gradient to certain internal convolutional layers. Despite this constraint, the summarized activations from the final convolutional layer proved informative for model tuning and qualitative assessment.
\vspace{0.25cm}

\noindent Third, reproducibility is constrained by contractual obligations: The dataset, model, and code used in this work are proprietary and cannot be made publicly available. 
However, we acknowledge the need for external validation and plan to benchmark the approach on public datasets in follow-up work. Practical constraints also prevented experimental comparisons with heavier video-specific explainability methods (3D attention maps, transformer-based attention visualizations), which typically require large-scale training and substantial GPU resources.
\vspace{0.25cm}

\noindent While a direct experimental comparison with other explainability pipelines was not feasible at this time, our visualizations and performance metrics establish a valuable baseline for on-going work.
Based on the above, we identify the following priorities for future research to:
\begin{itemize}
    \item Improve the reproducibility of our approach and validate our findings, it is essential to test it on public datasets. These tests will help demonstrate that the insights gained from visualization are a generalizable tool for hyperparameter tuning, and not just specific to our dataset;
    
    \item Improve contour and intensity detection in our visual outputs to better represent activation strength and reduce ambiguity in interpretation;
    
    \item Explore integration with lightweight object-detection modules to localize anomalies more precisely and reduce false positives, and evaluate the trade-off between added accuracy and inference cost;
    
    \item Investigate the integration of an attention mechanism into our architecture to compare its visual explanations with gradient-based methods and determine whether attention enhances our model interpretability;
    
    \item and extend visualization analysis using RISE (Randomized Input Sampling for Explanation) or its extensions tailored for time series, such as TimeREISE~\cite{timeREISE2022}. This approach can be applied not only to gradient and saliency maps but also to attention maps, offering a robust framework to evaluate model explainability during both training and evaluation phases.
\end{itemize}

\noindent In summary, this research work proposes a practical and transitional approach to improve the interpretability of TimeDistributed CNN+RNN models: By adapting established CNN visualization techniques to the TimeDistributed CNN+GRU setting, it provides concrete visual baselines and actionable tuning insights for practitioners. 
Situated within a fast-growing line of research on XAI, particularly for convolutional networks and image/video processing~\cite{olah2018the,olah2017feature,bau2020units,zhang2018interpretable} this study offers a pragmatic contribution that facilitates model interpretation and development.

\bibliographystyle{unsrtnat}
\bibliography{refs}

@INPROCEEDINGS{10479307,
  author={Poirier, Fabien and Jaziri, Rakia and Srour, Camille and Bernard, Gilles},
  booktitle={2023 20th ACS/IEEE International Conference on Computer Systems and Applications (AICCSA)}, 
  title={Enhancing Anomaly Detection in Videos using a Combined YOLO and a VGG GRU Approach}, 
  year={2023},
  volume={},
  number={},
  pages={1-6},
  doi={10.1109/AICCSA59173.2023.10479307}
}

@misc{poirier2024realtimeanomalydetectionvideo,
      title={Real-Time Anomaly Detection in Video Streams}, 
      author={Fabien Poirier},
      year={2024},
      eprint={2411.19731},
      archivePrefix={arXiv},
      primaryClass={cs.CV},
      url={https://arxiv.org/abs/2411.19731}, 
}

@book{algaba2019handbook,
  title={Handbook of the Shapley value},
  author={Algaba, Encarnaci{\'o}n and Fragnelli, Vito and S{\'a}nchez-Soriano, Joaqu{\'\i}n},
  year={2019},
  publisher={CRC Press}
}

@book{molnar2019, 
title = {Interpretable Machine Learning}, 
subtitle = {A Guide for Making Black Box Models Explainable},
author = {Christoph Molnar}, 
publisher = {Leanpub},
year = {2020}, 
pages =  {320}
}

@inproceedings{ribeiro2016should,
  title={" Why should i trust you?" Explaining the predictions of any classifier},
  author={Ribeiro, Marco Tulio and Singh, Sameer and Guestrin, Carlos},
  booktitle={Proceedings of the 22nd ACM SIGKDD international conference on knowledge discovery and data mining},
  pages={1135--1144},
  year={2016}
}

@article{lundberg2017unified,
  title={A unified approach to interpreting model predictions},
  author={Lundberg, Scott M and Lee, Su-In},
  journal={Advances in neural information processing systems},
  volume={30},
  year={2017}
}

@article{jiang2021layercam,
  title={Layercam: Exploring hierarchical class activation maps for localization},
  author={Jiang, Peng-Tao and Zhang, Chang-Bin and Hou, Qibin and Cheng, Ming-Ming and Wei, Yunchao},
  journal={IEEE Transactions on Image Processing},
  volume={30},
  pages={5875--5888},
  year={2021},
  publisher={IEEE}
}

@article{smilkov2017smoothgrad,
  title={Smoothgrad: removing noise by adding noise},
  author={Smilkov, Daniel and Thorat, Nikhil and Kim, Been and Vi{\'e}gas, Fernanda and Wattenberg, Martin},
  journal={arXiv preprint arXiv:1706.03825},
  year={2017}
}

@article{simonyan2014deep,
  title={Deep inside convolutional networks: Visualising image classification models and saliency maps},
  author={Simonyan, Karen and Vedaldi, Andrea and Zisserman, Andrew},
  booktitle={In Workshop at International Conference on Learning Representations},
  year={2014},
  journal={arXiv V2, arxiv.org:1312.6034}
}

@inproceedings{selvaraju2017grad,
  title={Grad-cam: Visual explanations from deep networks via gradient-based localization},
  author={Selvaraju, Ramprasaath R and Cogswell, Michael and Das, Abhishek and Vedantam, Ramakrishna and Parikh, Devi and Batra, Dhruv},
  booktitle={Proceedings of the IEEE international conference on computer vision},
  pages={618--626},
  year={2017}
}

@article{aditya1710grad,
  title={Grad-CAM++: Improved Visual Explanations for Deep Convolutional Networks},
  author={Aditya, C and Anirban, S and Abhishek, D and Prantik, H},
  journal={arXiv preprint arXiv:1710.11063},
  year={2018}
}

@article{bach2015pixel,
  title={On pixel-wise explanations for non-linear classifier decisions by layer-wise relevance propagation},
  author={Bach, Sebastian and Binder, Alexander and Montavon, Gr{\'e}goire and Klauschen, Frederick and M{\"u}ller, Klaus-Robert and Samek, Wojciech},
  journal={PloS one},
  volume={10},
  number={7},
  pages={e0130140},
  year={2015},
  publisher={Public Library of Science San Francisco, CA USA}
}

@inproceedings{sundararajan2017axiomatic,
  title={Axiomatic attribution for deep networks},
  author={Sundararajan, Mukund and Taly, Ankur and Yan, Qiqi},
  booktitle={International conference on machine learning},
  pages={3319--3328},
  year={2017},
  organization={PMLR}
}

@inproceedings{shrikumar2017learning,
  title={Learning important features through propagating activation differences},
  author={Shrikumar, Avanti and Greenside, Peyton and Kundaje, Anshul},
  booktitle={International conference on machine learning},
  pages={3145--3153},
  year={2017},
  organization={PMlR}
}

@article{springenberg2014striving,
  title={Striving for simplicity: The all convolutional net},
  author={Springenberg, Jost Tobias and Dosovitskiy, Alexey and Brox, Thomas and Riedmiller, Martin},
  journal={arXiv preprint arXiv:1412.6806},
  year={2014}
}

@article{englebert2024poly,
  title={Poly-cam: high resolution class activation map for convolutional neural networks},
  author={Englebert, Alexandre and Cornu, Olivier and Vleeschouwer, Christophe De},
  journal={Machine Vision and Applications},
  volume={35},
  number={4},
  pages={89},
  year={2024},
  publisher={Springer}
}

@inproceedings{muhammad2020eigen,
  title={Eigen-cam: Class activation map using principal components},
  author={Muhammad, Mohammed Bany and Yeasin, Mohammed},
  booktitle={2020 international joint conference on neural networks (IJCNN)},
  pages={1--7},
  year={2020},
  organization={IEEE}
}

@article{mundhenk2019efficient,
  title={Efficient saliency maps for explainable AI},
  author={Mundhenk, T Nathan and Chen, Barry Y and Friedland, Gerald},
  journal={arXiv preprint arXiv:1911.11293},
  year={2019}
}

@misc{petsiuk2018riserandomizedinputsampling,
      title={RISE: Randomized Input Sampling for Explanation of Black-box Models}, 
      author={Vitali Petsiuk and Abir Das and Kate Saenko},
      year={2018},
      eprint={1806.07421},
      archivePrefix={arXiv},
      primaryClass={cs.CV},
      url={ https://arxiv.org/abs/1806.07421}, 
}

@article{brocki2023class,
  title={Class-discriminative attention maps for vision transformers},
  author={Brocki, Lennart and Binda, Jakub and Chung, Neo Christopher},
  journal={arXiv preprint arXiv:2312.02364},
  year={2023}
}

@article{lozupone2024axial,
  title={AXIAL: Attention-based eXplainability for Interpretable Alzheimer's Localized Diagnosis using 2D CNNs on 3D MRI brain scans},
  author={Lozupone, Gabriele and Bria, Alessandro and Fontanella, Francesco and Meijer, Frederick JA and De Stefano, Claudio},
  journal={arXiv preprint arXiv:2407.02418},
  year={2024}
}

@misc{uchiyama2023adaptiveocclusionsensitivityanalysis,
      title={Adaptive occlusion sensitivity analysis for visually explaining video recognition networks}, 
      author={Tomoki Uchiyama and Naoya Sogi and Satoshi Iizuka and Koichiro Niinuma and Kazuhiro Fukui},
      year={2023},
      eprint={2207.12859},
      archivePrefix={arXiv},
      primaryClass={cs.CV},
      url={https://arxiv.org/abs/2207.12859}, 
}

@misc{zeiler2013visualizingunderstandingconvolutionalnetworks,
      title={Visualizing and Understanding Convolutional Networks}, 
      author={Matthew D Zeiler and Rob Fergus},
      year={2013},
      eprint={1311.2901},
      archivePrefix={arXiv},
      primaryClass={cs.CV},
      url={https://arxiv.org/abs/1311.2901}, 
}

@article{fernandez2024interpretability,
  title={Interpretability in video-based Human action recognition: saliency maps and GradCAM in 3D convolutional neural networks},
  author={Fernandez, Jorge Garcia-Torres and Meinich-Bache, {\O}yvind},
  journal={Nordic Machine Intelligence},
  volume={4},
  number={2},
  pages={19--25},
  year={2024}
}

@article{Fayyaz2025,
title = {Grad-CAM (Gradient-weighted Class Activation Mapping): A systematic literature review},
journal = {Computers in Biology and Medicine},
volume = {198},
pages = {111200},
year = {2025},
issn = {0010-4825},
doi = {https://doi.org/10.1016/j.compbiomed.2025.111200},
url = {https://www.sciencedirect.com/science/article/pii/S0010482525015537},
author = {Abdul Muiz Fayyaz and Said Jadid Abdulkadir and Noureen Talpur and Safwan Mahmood Al-Selwi and Shahab Ul Hassan and Ebrahim Hamid Sumiea}
}

@misc{raghakotkerasvis,
  title={keras-vis},
  author={Kotikalapudi, Raghavendra and contributors},
  year={2017},
  publisher={GitHub},
  howpublished={\url{https://github.com/raghakot/keras-vis}},
  note={Last access: August 2025.}
}

@misc{Keract,
  author = {Philippe Remy},
  title = {Keract: A library for visualizing activations and gradients},
  year = {2020},
  publisher = {GitHub},
  journal = {GitHub repository},
  howpublished = {\url{https://github.com/philipperemy/keract}},
  note={Last access: August 2025}
}

@article{gotkowski2020m3d,
  title={M3d-CAM: A PyTorch library to generate 3D data attention maps for medical deep learning},
  author={Gotkowski, Karol and Gonzalez, Camila and Bucher, Andreas and Mukhopadhyay, Anirban},
  journal={arXiv preprint arXiv:2007.00453},
  year={2020}
}

@misc{chollet2015keras,
 title={Keras},
 author={Chollet, Francois and others},
 year={2015},
 publisher={GitHub},
 howpublished={\url{https://github.com/fchollet/keras}},
 note={Last access: August 2025.}
}

@article{olah2018the,
  author = {Olah, Chris and Satyanarayan, Arvind and Johnson, Ian and Carter, Shan and Schubert, Ludwig and Ye, Katherine and Mordvintsev, Alexander},
  title = {The Building Blocks of Interpretability},
  journal = {Distill},
  year = {2018},
  note = {https://distill.pub/2018/building-blocks},
  doi = {10.23915/distill.00010}
}

@article{olah2017feature,
  author = {Olah, Chris and Mordvintsev, Alexander and Schubert, Ludwig},
  title = {Feature Visualization},
  journal = {Distill},
  year = {2017},
  note = {https://distill.pub/2017/feature-visualization},
  doi = {10.23915/distill.00007}
}

@article{bau2020units,
  author = {Bau, David and Zhu, Jun-Yan and Strobelt, Hendrik and Lapedriza, Agata and Zhou, Bolei and Torralba, Antonio},
  title = {Understanding the role of individual units in a deep neural network},
  elocation-id = {201907375},
  year = {2020},
  doi = {10.1073/pnas.1907375117},
  publisher = {National Academy of Sciences},
  issn = {0027-8424},
  URL = {https://www.pnas.org/content/early/2020/08/31/1907375117},
  journal = {Proceedings of the National Academy of Sciences}
}

@inproceedings{zhang2018interpretable,
  title={Interpretable convolutional neural networks},
  author={Zhang, Quanshi and Wu, Ying Nian and Zhu, Song-Chun},
  booktitle={Proceedings of the IEEE conference on computer vision and pattern recognition},
  pages={8827--8836},
  year={2018}
}

@inproceedings{gotkowski2021m3d,
  title={M3d-CAM: A PyTorch library to generate 3D attention maps for medical deep learning},
  author={Gotkowski, Karol and Gonzalez, Camila and Bucher, Andreas and Mukhopadhyay, Anirban},
  booktitle={Bildverarbeitung f{\"u}r die Medizin 2021: Proceedings, German Workshop on Medical Image Computing, Regensburg, March 7-9, 2021},
  pages={217--222},
  year={2021},
  organization={Springer}
}

@article{mahmoudi2023review,
  title={A review and comparative study of explainable deep learning models applied on action recognition in real time},
  author={Mahmoudi, Sidi Ahmed and Amel, Otmane and Stassin, Sedrick and Liagre, Margot and Benkedadra, Mohamed and Mancas, Matei},
  journal={Electronics},
  volume={12},
  number={9},
  pages={2027},
  year={2023},
  publisher={MDPI}
}

@article{roy2023explainable,
  title={Explainable activity recognition in videos using deep learning and tractable probabilistic models},
  author={Roy, Chiradeep and Nourani, Mahsan and Arya, Shivvrat and Shanbhag, Mahesh and Rahman, Tahrima and Ragan, Eric D and Ruozzi, Nicholas and Gogate, Vibhav},
  journal={ACM Transactions on Interactive Intelligent Systems},
  volume={13},
  number={4},
  pages={1--32},
  year={2023},
  publisher={ACM New York, NY}
}

@article{abdullah2024visioncam,
  title={Visioncam: A comprehensive XAI toolkit for interpreting image-based deep learning models},
  author={Abdullah, W and Tolba, A and Elmasry, A and Mostafa, NN},
  journal={Sustainable machine intelligence journal},
  volume={8},
  pages={4--46},
  year={2024}
}

@article{timeREISE2022,
AUTHOR = {Mercier, Dominique and Dengel, Andreas and Ahmed, Sheraz},
TITLE = {TimeREISE: Time Series Randomized Evolving Input Sample Explanation},
JOURNAL = {Sensors},
VOLUME = {22},
YEAR = {2022},
NUMBER = {11},
ARTICLE-NUMBER = {4084},
URL = {https://www.mdpi.com/1424-8220/22/11/4084},
PubMedID = {35684703},
ISSN = {1424-8220},
DOI = {10.3390/s22114084}
}

@misc{rgpdArticle22,
  author={GDPR —},
  title        = {Article 22: Automated individual decision-making, including profiling},
  howpublished = {\url{https://gdpr-info.eu/art-22-gdpr/}},
  note         = {Adopted on April 27, 2016, enforced on May 25, 2018, Last Access: September 25},
  year         = {2018}
}

@inproceedings{caron2021emerging,
  title={Emerging properties in self-supervised vision transformers},
  author={Caron, Mathilde and Touvron, Hugo and Misra, Ishan and J{\'e}gou, Herv{\'e} and Mairal, Julien and Bojanowski, Piotr and Joulin, Armand},
  booktitle={Proceedings of the IEEE/CVF international conference on computer vision},
  pages={9650--9660},
  year={2021}
}

\end{document}